\documentclass[10pt]{article}
\usepackage[preprint]{tmlr}

\usepackage[utf8]{inputenc}
\usepackage[T1]{fontenc}
\usepackage{hyperref}
\hypersetup{pdfauthor={Abhisek Keshari}, pdftitle={What Does a ProcGen Generalization Gap Measure? Action Rules, Residual Entropy, and the Missing Random Floor}, pdfsubject={}, pdfkeywords={}, pdfcreator={}}
\usepackage{url}
\usepackage{booktabs}
\usepackage{amsmath}
\usepackage{amssymb}
\usepackage{microtype}
\usepackage{graphicx}
\usepackage{placeins}
\usepackage{float}

\input{numbers.tex}
\input{numbers_carried.tex}
\InputIfFileExists{numbers_merged.tex}{}{}
\InputIfFileExists{numbers_cobbe.tex}{}{}

\title{What Does a ProcGen Generalization Gap Measure?\\
Action Rules, Residual Entropy, and the Missing Random Floor}

\author{\name Abhisek Keshari \email abhisek.keshari12@gmail.com \\
      \addr Independent researcher}

\begin{document}

\maketitle

\begin{abstract}
A generalization gap in reinforcement learning, return on training levels minus return on held-out levels, is usually reported without a reference point.
We argue that it should be read against a measured random floor: the return of a uniform-random policy on the same levels under the same evaluation harness.
On eight ProcGen environments with PPO at a compute-limited budget (8M steps, 16 parallel environments; three games extended to 25M), the floor changes what standard numbers mean.
The test-time action rule decides which policy is measured: in miner, the sampled policy scores \N{ratio.miner.8M.s.test}$\times$ the floor on held-out levels while its argmax scores below it in every run, and greedy evaluation places \N{count.below.8M.g.word} environments significantly below the floor.
Used as a convergence diagnostic, raw policy entropy flags \N{count.raw.8M.word} of eight environments, but \N{share.min.8M}--\N{share.max.8M}\% of that entropy lies on actions with identical effects; against the floor, \N{count.above.8M.s.word} of eight sampled policies are clearly above it on held-out levels and heist's is not distinguishable from it.
An audit of twelve ProcGen codebases finds that nine sample test-time actions with no explicit choice at the evaluation call site.
We recommend that every reported gap state its action rule, seed its evaluation and specify its tests before analysis, and report the floor on both level sets.

\end{abstract}

\section{Introduction}
\label{sec:intro}

A generalization gap in reinforcement learning (RL), return on training levels minus return on held-out levels, is the field's headline transfer metric \citep{cobbe2019quantifying, kirk2023survey}, and it is almost always reported without a reference point.\footnote{The author used AI assistants (Claude) for code implementation, analysis scripting and editing of the manuscript. The research questions, experimental design, analysis plan, interpretation and all claims are the author's; every reported number was produced by, and checked against, the analysis code.}
This paper argues that it should be read against a measured random floor: the return of a uniform-random policy on the same levels, under the same evaluation harness.
The floor costs minutes to measure, and on the benchmark we study it changes what several standard numbers mean.

The setting is ProcGen \citep{cobbe2020procgen}.
Each ProcGen environment procedurally generates \emph{levels} whose layout, textures and object placement vary under a seeded generator.
An agent is trained on a fixed set of levels (here 200) and evaluated on held-out levels from the same generator (here 100).
Algorithms, architectures and regularizers are compared by their gaps.
We study a compute-limited regime: PPO for 8M environment steps with 16 parallel environments, with three games extended to 25M steps (Section~\ref{sec:training}).

The first thing the floor exposes is that the test-time action rule decides which policy is measured.
An action can be \emph{sampled} from the policy's distribution or chosen greedily as its most probable action (argmax).
For a sharply peaked policy the two nearly coincide; otherwise they are different policies, and the difference is not a scalar inflation.
On identical checkpoints and identical levels, switching rules moves held-out return in both directions (Section~\ref{sec:results_protocol}).
In miner, sampling from the trained policy scores \N{ratio.miner.8M.s.test}$\times$ the floor on held-out levels, while its argmax scores below the floor in all \N{d.miner.8M.g.test.n} runs.
A policy that performs well above chance on held-out levels is measured as worse than chance.

The second is what policy entropy can and cannot say.
If raw policy entropy, against the maximum of $\ln 15 = 2.708$ over ProcGen's 15 discrete actions, is used as a convergence or ``policy has collapsed'' diagnostic, it mis-reads in identifiable ways. With cuts we set at 2.0 and 2.3 nats, it flags \N{count.raw.8M.word} of eight environments at our 8M-step budget as having intermediate or high residual entropy.
But ProcGen games have between 4 and 11 functionally distinct actions, and \N{share.min.8M}--\N{share.max.8M}\% of the measured entropy lies on actions with identical effects.
Read against the floor, the screen agrees at the extremes and over-calls in the middle: miner's policy is flagged by raw entropy and scores far above the floor in every run.
Heist is the clearest case at the other end: its sampled held-out return is not distinguishable from the floor ($\Delta$ = \N{d.heist.8M.s.test}) while longer training lifts its training-level return well above the floor (\N{heist.25M.s.train} against \N{floor.heist.train} at 25M) without moving its held-out return.
High entropy is not proof that training failed, and a stochastic policy is a legitimate object of evaluation.

Neither check is standard.
Auditing the released evaluation code of twelve prior ProcGen codebases, we find that all eleven that evaluate on held-out levels sample the test-time actions of their policy-gradient agents, nine with no explicit choice at the evaluation call site (Section~\ref{sec:audit}).
Sampling on both level sets is a matched protocol and measures the policy that was trained; what the audit documents is a rule that in most codebases is inherited rather than set, so its consequences for the reported gap go unexamined.

Our contributions:
\begin{enumerate}
  \item A measured random floor as the reference point for generalization gaps, taken under the evaluating harness on the same seeded level draws as the policy, on both level sets, for eight ProcGen environments at 8M and three at 25M.
  \item Evidence that the test-time action rule decides which policy is measured: on identical checkpoints and levels the two rules disagree in both directions, and greedy evaluation places \N{count.below.8M.g} environments significantly below the floor, including one whose sampled policy scores \N{ratio.miner.8M.s.test}$\times$ it.
  \item An analysis of raw policy entropy used as a convergence diagnostic, with cuts we set: effective action counts per game, merged-action entropy, and an above-floor check with pre-specified equivalence and multiplicity control, which together show where a residual-entropy flag is and is not informative.
  \item A twelve-codebase audit of ProcGen evaluation practice, verified against pinned commits to the file and line.
  \item A self-audit of our own case study: an encoder effect read against the floor, and two protocol choices in our own harness (the level-draw seed and the choice of test) that move a conclusion.
\end{enumerate}

\section{Related Work}
\label{sec:related}

\textbf{Reference points for returns.} Normalizing returns against a random policy is long established.
Atari's human-normalized score places a random-policy return at zero and a human return at one \citep{mnih2015human}, and \citet{machado2018revisiting} revisit ALE evaluation protocols (sticky actions, episode termination, what to report) because such choices change reported scores.
\citet{cobbe2020procgen} normalize ProcGen returns with fixed per-game constants, taking the minimum from a policy trained with observations masked out; on their easy-mode scale the uniform-random floor we measure lies below that minimum in most games (for example \N{cobbe.coinrun.8M.floor_test} in coinrun; Appendix~\ref{sec:app_cobbe}).
These references are published constants applied to \emph{returns}.
Ours differs in three ways that matter for a gap: it is \emph{measured} under the same harness and episode accounting as the policy, on the \emph{same seeded level draws}, and on \emph{both} level sets, so it is subtracted from the two terms of a train-minus-test difference rather than rescaling one return.

\textbf{Evaluation methodology in RL.} \citet{henderson2018matters} document reproducibility failures from underspecified evaluation; \citet{jordan2020evaluating} propose performance measures and confidence intervals that account for algorithm and environment variability; \citet{agarwal2021precipice} argue for interquartile means and stratified bootstrap intervals at small seed counts; and \citet{patterson2024empirical} set out empirical-design practice for RL, including baselines, hypothesis testing and multiple comparisons.
Those works ask whether results are reported rigorously.
We ask a question one step earlier: whether the two numbers being differenced mean what they appear to, which requires knowing which policy produced them and where chance lies.

\textbf{Why the default exists.} Prior ProcGen protocols were designed largely for ranking.
The original benchmark \citep{cobbe2020procgen} and the NeurIPS 2020 ProcGen competition \citep{mohanty2021competition} average many episodes for a stable leaderboard, a setting in which stochastic action selection is internally consistent across submissions.
When a single train-minus-test difference becomes a paper's headline number, which rule produced it, whether it describes one checkpoint, and where it sits relative to chance all bear on what it means.

\textbf{Policy entropy and stagnation.} \citet{beukman2026stagnation} document that PPO frequently stagnates short of a task-solving policy, and several PPO variants address this \citep{rahman2022rpo, lixandru2024axppo, rahman2025ppobr, xie2025spo}.
\citet{juliani2024plasticity} show that on-policy agents lose plasticity under distribution shift, including on ProcGen, which is a further reason a policy can stall with high residual entropy.
Our increment concerns measurement: how residual entropy interacts with the action rule and with the floor when a gap is read.
Stochastic action at test time can also be a deliberate strength: \citet{zisselman2023explore} improve ProcGen generalization by taking exploratory actions at test time, so a sampled policy is a legitimate object of evaluation.

\textbf{Encoders.} The IMPALA CNN \citep{espeholt2018impala} is the standard ProcGen encoder; feature pyramid networks \citep{lin2017fpn} are a multi-scale design from object detection, and VSOP-3D \citep{jesson2024vsop3d} reports ProcGen gains from combining architectural scale-up with an algorithmic change \citep{jesson2024relu}.

\section{Method}
\label{sec:method}

\subsection{Agents}
\label{sec:training}

Two encoders share a three-block IMPALA backbone; the baseline flattens the final feature map and projects it to 256 dimensions, and the \emph{multiscale FPN} variant additionally fuses the three blocks' feature maps top-down before the same projection (details in Appendix~\ref{sec:app_extra}).
Both agents output a categorical (softmax) policy over the 15 discrete actions; heads, algorithm and hyperparameters are identical.
We train PPO on 200 training levels, easy mode, with CleanRL's ProcGen hyperparameters (Appendix~\ref{sec:app_extra}) but without its reward normalization and reward clipping, which our trainer does not apply; held-out evaluation uses a fixed set of 100 levels. Appendix~\ref{sec:app_cobbe} reports the resulting returns as normalized scores on the scale of \citet{cobbe2020procgen}.
The 8M-step runs cover eight environments with three seeds per encoder and use 16 parallel environments; runs extended to 25M steps (starpilot, heist, fruitbot; 9, 6 and 3 seeds per encoder) use 256 (the CleanRL default is 64).

\subsection{Evaluation protocol}
\label{sec:protocol}

Three properties of an evaluation protocol determine what a reported gap measures.

\textbf{Which policy.} Sampling evaluates the policy that was trained.
Argmax evaluates a deterministic policy derived from it, which coincides with the trained policy only when the action distribution is sharply peaked.
A harness can describe itself as deterministic while sampling at every step, so the rule has to be checked in code, not read from a configuration name.

\textbf{Which checkpoint.} A commonly logged metric is the running mean return of recent episodes collected during training.
Such a number averages over several policy versions and describes no single checkpoint.

\textbf{Relative to what.} We measure a uniform-random floor over the 15 actions with the same harness, level draws and episode accounting as the policies.

Every return in this paper comes from one protocol.
Each final checkpoint is evaluated under both rules on both level sets (200 training levels; 100 held-out levels starting at level 1000) on three seeded level draws, 128 episodes per draw (384 per run, rule and level set), with balanced per-slot episode counting and an episode burn-in.
A run's value is its mean over the three draws.
The floor is measured on exactly the same three draws, so every policy--floor comparison, at both budgets, is on identical levels.
Sampling is the primary rule throughout; greedy values are reported beside it.

\subsection{Entropy screen and above-floor check}
\label{sec:convergence}

We examine raw policy entropy as it would be used as a convergence or ``policy has collapsed'' diagnostic; the cuts below are ours. The \emph{screen} is policy entropy over raw actions at the final checkpoint, on the states visited by the sampled policy during evaluation, averaged over the two level sets.
It has three tiers: \textbf{low residual entropy} (below $2.0$ nats), \textbf{intermediate} ($[2.0, 2.3]$) and \textbf{high} (above $2.3$).
A policy in the low tier is what an entropy-based convergence criterion would call converged; we use the word only in that sense.
The \emph{above-floor check} asks whether sampled held-out return is significantly above the floor (Section~\ref{sec:stats}); it needs no threshold, normalizer or action-space assumption.
Entropy alone cannot deliver a verdict: a high-entropy policy may be a legitimately stochastic policy, and a low-entropy policy may still perform poorly. PPO's entropy bonus (coefficient 0.01) also keeps raw entropy elevated by design, one more reason a raw-entropy tier cannot by itself indicate a failure to learn.

\subsection{Statistical analysis}
\label{sec:stats}

After an initial single-draw analysis of the same 8M checkpoints, we wrote an analysis plan fixing the tests, families, margins and stop conditions, then ran the three-draw re-analysis reported here (which includes the original draw) under it. We call this a pre-specified confirmatory re-analysis rather than a pre-registration, because the plan was written after the initial results had been seen.
A floor comparison uses the difference $\Delta$ between the mean over runs and the floor, a two-sided Welch $t$-test combining the run-to-run SEM with the floor's per-episode SEM, and a 95\% percentile bootstrap CI ($10^4$ resamples).
We label a policy \emph{above} or \emph{below} the floor when the Holm-adjusted $p < 0.05$; otherwise \emph{equivalent to the floor} if a two one-sided test (TOST) with margin $\delta = \max(0.1\,|\text{floor}|, 0.25)$ passes, and \emph{not distinguishable} otherwise. In tables, the ``call'' is this label, and $k/n$ is the number of runs whose value lies above the floor (descriptive).
Holm's correction is applied within each family of eight environments (per rule, level set and budget), and within each family of gap tests.
Gaps are tested per run with Welch's $t$ on episode returns, and per-run $p$-values are combined across runs with Fisher's method, with Stouffer's $Z$ on one-sided $p$-values as a directional check; the three draws of a run share weights and are never combined as independent tests.

\section{Field-Wide Audit}
\label{sec:audit}

We audited the released evaluation code of twelve prior ProcGen codebases spanning ICML, NeurIPS, RLC and arXiv from 2019 to 2025.
For each repository we located the code path that produces the reported test numbers and classified two properties: (i)~\emph{stochastic evaluation}, whether test-time actions are sampled from the policy rather than chosen by argmax; and (ii)~\emph{in-loop test returns}, whether reported test returns come from a held-out rollout that runs inside the training loop and is reported as a running mean over recent episodes, which span several policy versions, rather than from an evaluation of a fixed checkpoint.
In every codebase we mark in-loop, the held-out environments or workers contribute no gradient.
Classifications were made against pinned commits, with the evidencing file and line for every entry in Appendix~\ref{sec:app_audit}.

\textbf{What we found.} Eleven of twelve codebases perform held-out evaluation; \emph{all eleven} evaluate their policy-gradient agents stochastically, and six of them report in-loop test returns (Table~\ref{tab:audit}).
UCB-DrAC \citep{raileanu2021ucbdrac} is the most explicit instance, passing \texttt{deterministic=False} at the call site.
Impoola \citep{trumpp2025impoola} sets the rule through a documented configuration option: its PPO evaluation samples and its DQN evaluation is greedy.
EDE \citep{jiang2023ede} is a value-based method; the row classifies its PPO baseline, which samples, while EDE's own agent is evaluated $\epsilon$-greedily ($\epsilon = 0.05$).
For a value-based agent the greedy action is the trained policy, so neither case bears on the sampled-versus-argmax analysis below.
In mixreg, the ProcGen baseline and IBAC-SNI the in-loop test numbers come from a dedicated test worker whose gradient is weighted to zero; in RAD, VSOP and VSOP-3D from a separate held-out environment whose transitions are discarded.
In all six the train number comes from the matching in-loop procedure on training levels, so train and test are measured the same way; the test number is a sampled-action running average taken during training rather than a measurement of the final policy.
PPG \citep{cobbe2021ppg} trains on the full level distribution and performs no held-out evaluation in its released code.

\textbf{Provenance.} Of the eleven, nine have no explicit choice at the evaluation call site.
Three (DAAC, PLR and EDE's PPO baseline) inherit sampling through a default argument that no call site overrides, and the six in-loop codebases sample through the training rollout's own sampling call.
The two exceptions show that the rule can be set deliberately: UCB-DrAC passes it at the call site, and Impoola exposes it as a configuration option.
By Section~\ref{sec:results_protocol}, sampling is the better of the two rules for policies with high residual entropy; the concern is that where it is inherited, its consequences for the reported gap go unexamined.

\textbf{Scope.} We audited released evaluation code, not results as reported in papers, and we do not claim that any published number is wrong.
We did not measure how in-loop averages compare with fixed-checkpoint evaluation in these codebases.

\begin{table}[t]
  \caption{Code audit of twelve prior ProcGen codebases. Stoch.: test-time actions of the policy-gradient agent are sampled. In-loop: test returns come from a held-out rollout inside the training loop, reported as a running mean over episodes that span policy updates (in none of the six is the policy trained on that rollout). $^*$UCB-DrAC passes \texttt{deterministic=False} explicitly. $^\dagger$Impoola sets the rule by configuration (PPO samples, DQN greedy). $^\ddagger$EDE's PPO baseline; EDE's own agent is evaluated $\epsilon$-greedily. Commits and file/line evidence: Appendix~\ref{sec:app_audit}.}
  \label{tab:audit}
  \centering
  \footnotesize
  \setlength{\tabcolsep}{4pt}
  \begin{tabular}{lccc}
    \toprule
    Codebase & Venue & Stoch. & In-loop \\
    \midrule
    VSOP-3D \citep{jesson2024vsop3d}    & arXiv'24   & \checkmark     & \checkmark \\
    VSOP \citep{jesson2024relu}         & ICML'24    & \checkmark     & \checkmark \\
    ProcGen \citep{cobbe2020procgen}    & ICML'20    & \checkmark     & \checkmark \\
    mixreg \citep{wang2020mixreg}       & NeurIPS'20 & \checkmark     & \checkmark \\
    RAD \citep{laskin2020rad}           & NeurIPS'20 & \checkmark     & \checkmark \\
    IBAC-SNI \citep{igl2019ibacsni}     & NeurIPS'19 & \checkmark     & \checkmark \\
    DAAC \citep{raileanu2021daac}       & ICML'21    & \checkmark     & -- \\
    UCB-DrAC \citep{raileanu2021ucbdrac}& NeurIPS'21 & \checkmark$^*$ & -- \\
    PLR \citep{jiang2021plr}            & ICML'21    & \checkmark     & -- \\
    EDE \citep{jiang2023ede}            & NeurIPS'23 & \checkmark$^\ddagger$ & -- \\
    Impoola \citep{trumpp2025impoola}   & RLC'25     & \checkmark$^\dagger$ & -- \\
    PPG \citep{cobbe2021ppg}            & ICML'21    & n/a            & n/a \\
    \bottomrule
  \end{tabular}
\end{table}

\section{Results}
\label{sec:results}

\subsection{The action rule decides which policy is measured}
\label{sec:results_protocol}

Table~\ref{tab:unmask} evaluates the 48 checkpoints at 8M steps (eight environments $\times$ two encoders $\times$ three seeds) on held-out levels under both rules, beside the floor on the same draws.

\textbf{One checkpoint, two answers.} Sampling from the trained miner policies scores \N{miner.8M.s.test} on held-out levels, \N{ratio.miner.8M.s.test}$\times$ the floor of \N{floor.miner.test}, above it in all \N{d.miner.8M.s.test.n} runs (three seeds $\times$ two encoders); the argmax of the same weights scores \N{miner.8M.g.test}, below the floor in all \N{d.miner.8M.g.test.n} runs (Holm $p$ = \N{d.miner.8M.g.test.ph}).
The sampled policy performs well above chance on held-out levels, and greedy evaluation reports it as worse than chance.
Coinrun shows the same pattern more weakly: sampled \N{coinrun.8M.s.test}, above its floor of \N{floor.coinrun.test}; greedy \N{coinrun.8M.g.test}, not distinguishable from it.

\textbf{The rule's effect has no fixed sign.} Paired over runs, sampled held-out return exceeds greedy in miner, coinrun and heist (by \N{rd.heist.8M.test.u} in heist, greedy lower in all six runs).
Greedy exceeds sampled in fruitbot (all six runs) and bossfight (\N{rd.bossfight.8M.test.kg} of six runs).
Starpilot, which has the lowest entropy, bigfish and dodgeball (sampled \N{dodgeball.8M.s.test} against greedy \N{dodgeball.8M.g.test}, paired $p$ = \N{rd.dodgeball.8M.test.p}) are within noise.
No single correction factor converts one rule's numbers into the other's.

\textbf{Why argmax can fall below chance.} Argmax of a policy with high residual entropy commits every state to whichever action narrowly leads, and where that preference is arbitrary the committed behavior can be worse than random, for example by repeating a failing trajectory where random play would eventually reach the goal (a plausible mechanism that we do not isolate).
A second candidate is specific to ProcGen's action space: argmax over raw actions can choose against the policy's own preference when one behavior's probability is split across functionally equivalent actions (Appendix~\ref{sec:app_calib}); in miner, the three leftward and three rightward moves each collapse to one.

\textbf{An exploratory check on the mechanism.} In an analysis outside the original plan, whose rule and reading were fixed in an addendum before it was run (Appendix~\ref{sec:app_merged}; own Holm family), we removed the second candidate while keeping a deterministic policy: sum the policy's probabilities within each per-game class of equivalent actions, take the top class, then its most probable action. This merged argmax changes the chosen class on \N{m.miner.test.dis}\%, \N{m.heist.test.dis}\% and \N{m.coinrun.test.dis}\% of held-out steps in miner, heist and coinrun, yet it does not lift them: held-out return stays below the floor in miner ($\Delta$ = \N{m.miner.test.d}) and heist ($\Delta$ = \N{m.heist.test.d}), and in coinrun it is not distinguishable from the floor and lower than plain argmax (\N{m.coinrun.test.mg}, $p$ = \N{m.coinrun.test.mg.p}). Across the eight environments it never significantly exceeds plain argmax and is lower in five. This is consistent with the first candidate rather than the second; it does not isolate the mechanism, because the per-game partition leaves actions that are equivalent only in some states unmerged. Merged argmax also lowers return in starpilot (\N{m.starpilot.test.mg}, $p$ = \N{m.starpilot.test.mg.p} at six runs), where the two rules should nearly coincide, which suggests the per-game classes, measured on random-play states, may merge actions that differ on states the trained policy visits; the check therefore bounds the second candidate only loosely.

\textbf{When the sampled number is near chance.} Heist's sampled held-out return (\N{heist.8M.s.test}) is not distinguishable from the floor (\N{floor.heist.test}; $\Delta$ = \N{d.heist.8M.s.test}, 95\% CI [\N{d.heist.8M.s.test.lo}, \N{d.heist.8M.s.test.hi}]), and equivalence within the pre-specified margin is not established either (TOST $p$ = \N{d.heist.8M.s.test.ptost}).
Heist pays 0 or 10 per level, so the sampled policy completes about \N{succ.heist.8M.s.test}\% of held-out levels, random play about \N{succ.heist.floor.test}\% and the greedy policy about \N{succ.heist.8M.g.test}\%.
Read as a ratio of sampled to greedy it looks like stochastic inflation; read against the floor it is greedy deflation below chance.

\textbf{What this does to the gap.} One rule applied to both level sets is a matched measurement, and the audit found no codebase that takes train and test numbers from different procedures.
The concern is that the rule decides which policy is measured, so the gaps differ too.
At 8M the sampled gap exceeds the greedy gap in heist (\N{gap.heist.8M.s.u} against \N{gap.heist.8M.g.u}; paired difference \N{gd.heist.8M} [\N{gd.heist.8M.lo}, \N{gd.heist.8M.hi}], $p$ = \N{gd.heist.8M.p}) and miner (\N{gap.miner.8M.s.u} against \N{gap.miner.8M.g.u}), and is within noise elsewhere.
Under sampling, heist's training-level return (\N{heist.8M.s.train}) is above its floor (\N{floor.heist.train}) by \N{d.heist.8M.s.train.u} but not significantly after correction (Holm $p$ = \N{d.heist.8M.s.train.ph}), while its held-out return is not distinguishable from the floor.
Under greedy selection both levels sit below the floor: the greedy gap is a difference between two sub-chance numbers.

\begin{table}[t]
  \caption{Held-out return of the same 8M checkpoints under sampled (primary) and greedy action selection, against a uniform-random floor on identical level draws. Returns are mean $\pm$ SEM over six runs (two encoders $\times$ three seeds; each run averaged over three seeded draws of 128 episodes); the floor is mean $\pm$ per-episode SEM over the same three draws. $\Delta$ is sampled return minus floor with a 95\% bootstrap CI. Call (primary, pre-specified): above/below if Holm-adjusted Welch $p<0.05$, equiv.\ if TOST passes, n.d.\ (not distinguishable) otherwise. Descriptive columns: $z$ for a normal-approximation rule ($z > 2$; $\Delta$ over the combined SEM) and the number of runs above the floor, $k$/6.}
  \label{tab:unmask}
  \centering
  \small
  \setlength{\tabcolsep}{3.5pt}
  \begin{tabular}{lrrrlrrrlrr}
\toprule
 & & \multicolumn{5}{c}{Sampled (primary)} & \multicolumn{4}{c}{Greedy} \\
\cmidrule(lr){3-7}\cmidrule(lr){8-11}
Env & Floor & Return & $\Delta$ [95\% CI] & Call & $z$ & $k$/6 & Return & Call & $z$ & $k$/6 \\
\midrule
starpilot & $1.67 \pm 0.11$ & $18.35 \pm 1.11$ & $+16.68$ $[14.78, 18.72]$ & above & $+15.0$ & 6 & $17.89 \pm 1.40$ & above & $+11.6$ & 6 \\
fruitbot & $-2.66 \pm 0.21$ & $22.89 \pm 0.39$ & $+25.55$ $[24.72, 26.34]$ & above & $+58.0$ & 6 & $25.79 \pm 0.41$ & above & $+61.7$ & 6 \\
bigfish & $0.84 \pm 0.06$ & $2.82 \pm 0.75$ & $+1.98$ $[0.72, 3.37]$ & n.d. & $+2.6$ & 6 & $3.19 \pm 0.89$ & n.d. & $+2.6$ & 6 \\
coinrun & $3.28 \pm 0.24$ & $5.72 \pm 0.31$ & $+2.44$ $[1.73, 3.17]$ & above & $+6.3$ & 6 & $3.57 \pm 0.62$ & n.d. & $+0.4$ & 3 \\
miner & $1.17 \pm 0.10$ & $5.93 \pm 0.30$ & $+4.76$ $[4.19, 5.33]$ & above & $+14.9$ & 6 & $0.60 \pm 0.07$ & below & $-4.6$ & 0 \\
dodgeball & $0.47 \pm 0.05$ & $1.09 \pm 0.13$ & $+0.62$ $[0.40, 0.92]$ & above & $+4.3$ & 6 & $0.82 \pm 0.05$ & above & $+4.9$ & 6 \\
bossfight & $0.09 \pm 0.05$ & $1.36 \pm 0.49$ & $+1.27$ $[0.44, 2.17]$ & n.d. & $+2.6$ & 6 & $2.28 \pm 0.69$ & n.d. & $+3.2$ & 6 \\
heist & $3.15 \pm 0.24$ & $2.88 \pm 0.21$ & $-0.27$ $[-0.86, 0.31]$ & n.d. & $-0.9$ & 2 & $0.45 \pm 0.14$ & below & $-9.8$ & 0 \\
\bottomrule
\end{tabular}

\end{table}

\subsection{The entropy screen against the above-floor check}
\label{sec:results_convergence}

Table~\ref{tab:tiers} sets the screen beside the above-floor check at 8M.
By raw entropy, starpilot and bigfish have low residual entropy, fruitbot, coinrun and miner intermediate, and dodgeball, bossfight and heist high.
Against the floor, \N{count.above.8M.s} of eight sampled policies are above it on held-out levels (starpilot, fruitbot, coinrun, miner, dodgeball); bigfish and bossfight are inconclusive at six runs (their $z$ is near \N{d.bigfish.8M.s.test.z}, but neither survives Holm correction, and each splits by encoder); heist is not distinguishable from it.

The screen and the check agree at the extremes: starpilot is low and above the floor, heist is high and not above it.
In the middle the screen over-calls.
Miner is the clearest case: flagged by raw entropy (\N{H.miner.8M} nats, \N{H.miner.8M.pct}\% of maximum), its sampled policy is above the floor in all six runs.
Fruitbot, also flagged (intermediate tier), is likewise far above the floor (Holm $p$ = \N{d.fruitbot.8M.s.test.ph}).
A raw-entropy flag says that the distribution over raw actions is broad; it does not by itself show that the policy performs at chance.

Part of the over-call is the action space.
Entropy over 15 raw actions counts probability spread across actions with identical effects as indecision, and \N{share.min.8M}--\N{share.max.8M}\% of each environment's raw entropy is of that kind (Appendix~\ref{sec:app_calib}).
After summing the probabilities of equivalent actions, \N{count.state.8M} environments remain outside the low tier when equivalence is determined state by state (coinrun and heist) and \N{count.game.8M} under one partition per game (dodgeball, bossfight, heist); these counts depend on expressing the cuts as the same fractions of the maximum, which is our choice.
We keep raw-action entropy for the tiers because it is what practitioners log, and read an intermediate or high tier as a prompt to run the above-floor check.

Two robustness checks bound the tiers.
Sweeping the low/intermediate cut across $[1.8, 2.2]$ leaves starpilot low and heist high at every value, while the environments near 2.0 move with the cut (Appendix~\ref{sec:app_threshold}); since starpilot and heist are the extremes of the observed range, their stability follows from the data layout rather than from robustness.
Measuring entropy on greedy-evaluation states instead moves two environments' raw tier by one step, bigfish up and miner down; measuring it only on the first 1024 evaluation states, a short window, moves values substantially in both directions (Table~\ref{tab:merged}: starpilot at 25M, \N{H.starpilot.25M.first} against \N{H.starpilot.25M.probe} on a uniform sample of all visited states; miner at 8M, \N{H.miner.8M.first} against \N{H.miner.8M.probe}).

\begin{table}[t]
  \caption{Entropy screen and above-floor check at 8M, pooled over both encoders and three seeds. $H$: mean raw-action entropy on sampled-evaluation states, averaged over train and test levels (maximum $\ln 15 = 2.708$). Tiers: low ($<2.0$), intermediate ($[2.0, 2.3]$), high ($>2.3$) residual entropy; merged tiers after summing the probabilities of functionally equivalent actions per state / per game, with cuts at the same fractions of the maximum. Share: fraction of raw entropy on equivalent actions. Above-floor: primary call of Table~\ref{tab:unmask}. Gaps: pooled train$-$test return under sampled (s) and greedy (g) selection.}
  \label{tab:tiers}
  \centering
  \small
  \setlength{\tabcolsep}{3.5pt}
  \begin{tabular}{lrrlllrcr}
\toprule
Env & $H$ & \% max & Raw tier & Merged (state / game) & Share & Above-floor & Gap (s) & Gap (g) \\
\midrule
starpilot & $1.794$ & $66.2$ & low & low / low & $37\%$ & above & $-0.23$ & $-1.26$ \\
fruitbot & $2.078$ & $76.7$ & interm. & low / low & $66\%$ & above & $+1.22$ & $+0.88$ \\
bigfish & $1.971$ & $72.8$ & low & low / low & $33\%$ & n.d. & $+1.58$ & $+1.43$ \\
coinrun & $2.261$ & $83.5$ & interm. & interm. / low & $59\%$ & above & $+0.02$ & $+0.19$ \\
miner & $2.087$ & $77.1$ & interm. & low / low & $66\%$ & above & $+1.55$ & $+0.61$ \\
dodgeball & $2.427$ & $89.6$ & high & low / interm. & $32\%$ & above & $+0.59$ & $+0.46$ \\
bossfight & $2.329$ & $86.0$ & high & low / interm. & $41\%$ & n.d. & $+0.34$ & $+0.49$ \\
heist & $2.569$ & $94.9$ & high & interm. / interm. & $34\%$ & n.d. & $+1.54$ & $+0.71$ \\
\bottomrule
\end{tabular}

\end{table}

\subsection{Longer training under a different configuration}
\label{sec:results_budget}

\begin{figure}[t]
  \centering
  \includegraphics[width=\textwidth]{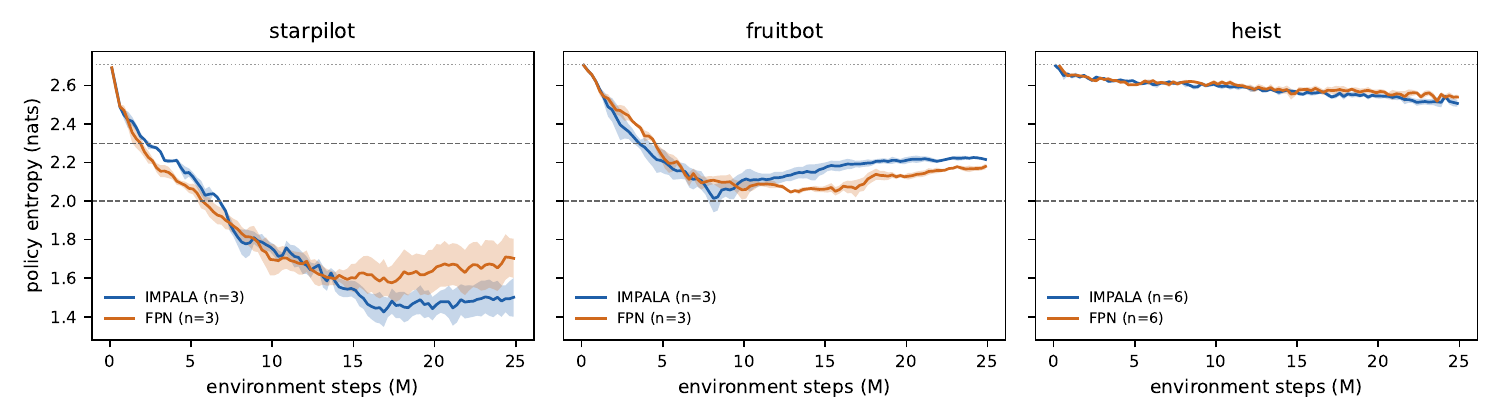}
  \caption{Raw-action policy entropy during training at 25M steps, from per-iteration training logs (sampled actions on training levels; mean over seeds per encoder, band $\pm 1$ SEM; $n$ in legend). Dashed lines: the screen's 2.0 and 2.3 cuts; dotted: $\ln 15$.}
  \label{fig:entropy25m}
\end{figure}

We extended starpilot, fruitbot and heist to 25M steps (Figure~\ref{fig:entropy25m}, Table~\ref{tab:res25m}).
These runs also used 256 parallel environments rather than 16, so they change batch size and the number of policy updates (about 381 against 1{,}953) along with the step budget; the comparison is between two configurations, not two training lengths.
Starpilot's raw entropy falls well below the low/intermediate cut and keeps descending; fruitbot's plateaus in the intermediate tier, largely over equivalent actions (\N{H.fruitbot.8M.share}\% of its raw entropy at 8M), while its policy is far above the floor.
Heist's entropy declines only slightly.
Under sampling on the floor's own level draws, heist's training-level return at 25M rises to \N{heist.25M.s.train} against a training floor of \N{floor.heist.train} (Holm $p$ = \N{d.heist.25M.s.train.ph}), while its held-out return is \N{heist.25M.s.test} against \N{floor.heist.test} ($\Delta$ = \N{d.heist.25M.s.test} [\N{d.heist.25M.s.test.lo}, \N{d.heist.25M.s.test.hi}]; \N{d.heist.25M.s.test.label}).
In this configuration the additional steps raise training-level return without a measurable held-out gain, so a longer run alone does not settle whether a heist gap reflects transfer; the above-floor check is what does.
The heist sampled gap at 25M (\N{gap.heist.25M.s.u}) is again larger than the greedy gap (\N{gap.heist.25M.g.u}; paired difference \N{gd.heist.25M} [\N{gd.heist.25M.lo}, \N{gd.heist.25M.hi}]).

A low tier does not make a gap stable across configurations.
Starpilot's sampled gap is \N{gap.starpilot.8M.s} [\N{gap.starpilot.8M.s.lo}, \N{gap.starpilot.8M.s.hi}] at 8M, indistinguishable from zero, and \N{gap.starpilot.25M.s} [\N{gap.starpilot.25M.s.lo}, \N{gap.starpilot.25M.s.hi}] at 25M (Appendix~\ref{sec:app_pooling}); the two configurations differ in parallelism as well as length.

\begin{table}[t]
  \caption{Final-checkpoint results at 25M steps, mean $\pm$ SEM over seeds (starpilot $n{=}9$, heist $n{=}6$, fruitbot $n{=}3$ per encoder); each run averaged over three seeded draws of 128 episodes per level set. Sampled is primary. Tier: raw-entropy screen on sampled-evaluation states.}
  \label{tab:res25m}
  \centering
  \small
  \setlength{\tabcolsep}{3.5pt}
  \begin{tabular}{llrrrrrl}
\toprule
 & & \multicolumn{3}{c}{Sampled (primary)} & \multicolumn{2}{c}{Greedy} & \\
\cmidrule(lr){3-5}\cmidrule(lr){6-7}
Env & Encoder & Train & Test & Gap & Test & Gap & Tier \\
\midrule
starpilot & IMPALA & $31.04 \pm 1.37$ & $28.02 \pm 1.55$ & $3.02 \pm 0.33$ & $30.15 \pm 1.97$ & $2.55 \pm 1.00$ & low \\
starpilot & FPN & $28.92 \pm 1.51$ & $27.72 \pm 1.73$ & $1.20 \pm 0.51$ & $29.08 \pm 1.97$ & $2.03 \pm 0.81$ & low \\
\addlinespace
fruitbot & IMPALA & $27.57 \pm 0.47$ & $21.39 \pm 0.62$ & $6.18 \pm 0.75$ & $23.79 \pm 0.69$ & $5.12 \pm 0.82$ & interm. \\
fruitbot & FPN & $26.19 \pm 0.13$ & $20.52 \pm 0.36$ & $5.66 \pm 0.49$ & $21.79 \pm 0.34$ & $5.80 \pm 0.52$ & interm. \\
\addlinespace
heist & IMPALA & $7.14 \pm 0.23$ & $2.66 \pm 0.26$ & $4.47 \pm 0.17$ & $0.62 \pm 0.15$ & $3.31 \pm 0.19$ & high \\
heist & FPN & $6.75 \pm 0.15$ & $3.34 \pm 0.25$ & $3.41 \pm 0.10$ & $1.22 \pm 0.14$ & $1.94 \pm 0.26$ & high \\
\addlinespace
\bottomrule
\end{tabular}

\end{table}

\subsection{Protocol choices in our own harness}
\label{sec:results_selfaudit}

The matched-protocol requirement extends to places we did not initially look.
ProcGen draws the evaluated subset of levels from an unseeded generator unless told otherwise, so re-evaluating identical checkpoints instantiates a different subset each time.
The three seeded draws used here show how much that matters: on heist at 25M, the sampled encoder difference in held-out return has $p$ between \N{encpd.heist.25M.s.test.pmin} and \N{encpd.heist.25M.s.test.pmax} across the three draws taken one at a time, crossing the conventional boundary, while the gap difference stays significant on every draw ($p \le$ \N{encpd.heist.25M.s.gap.pmax}).
We therefore seed every draw, average each run over three, and never combine re-evaluations of the same weights as independent tests.

A second choice is the test itself.
By a normal-approximation rule ($z > 2$), greedy evaluation leaves \N{count.v1atbelow.8M.g} environments at or below the floor at 8M; under the pre-specified Welch test with Holm correction it leaves \N{count.below.8M.g} significantly below and makes bigfish and bossfight inconclusive.
The two tests agree on every environment whose result carries the argument (miner and heist below under greedy; miner above under sampling), and they disagree on the moderate, encoder-dependent effects.
The plan's test was chosen after the single-draw counts had been seen, it changed one of them, and we report the plan's result. An evaluation protocol is not fully specified until its random draws and its tests are fixed before the analysis that reports them.

\subsection{Encoder comparison, read against the floor}
\label{sec:results_encoder}

On heist at 25M, IMPALA's gap exceeds FPN's under both rules, by \N{encd.heist.25M.g.gap.u} with greedy selection ($p$ = \N{encd.heist.25M.g.gap.p}) and by \N{encd.heist.25M.s.gap.u} with sampling ($p$ = \N{encd.heist.25M.s.gap.p}), so the effect survives the change of rule.
The floor says what it compares: the greedy held-out returns of both encoders are below the floor, and the sampled ones sit near it (Table~\ref{tab:res25m}; $z$ = \N{enc.heist.25M.s.test.impala.z} for IMPALA and \N{enc.heist.25M.s.test.fpn.z} for FPN), with \N{encd.heist.25M.s.share}\% of the sampled gap difference coming from training-level return and the rest from FPN's higher held-out return.
A significant gap difference between two policies whose held-out returns are near chance is a real measurement and weak evidence about generalization.
On starpilot at 25M the sampled encoder gap difference ($p$ = \N{encd.starpilot.25M.s.gap.p}) comes entirely from training-level return, with held-out returns not differing ($p$ = \N{encd.starpilot.25M.s.test.p}).
A within-encoder representation statistic we computed proved invalid on disjoint inputs (Appendix~\ref{sec:app_cka}).

\section{Discussion}
\label{sec:discussion}

\textbf{Report the rule and the floor.} The recommendation is not ``use sampling''; it is to state the action rule and report a measured floor on both level sets.
Sampling evaluates the policy that was trained, argmax a deterministic policy derived from it.
The two agree when residual entropy is low (starpilot's rules are within noise), and diverge exactly where it is not: every greedy floor crossing in Table~\ref{tab:unmask} occurs in an environment the screen flags.
This is a comparison across environments, so entropy is confounded with game identity; a within-game test would vary residual entropy in one environment.
Argmax remains reasonable when the deterministic policy is itself the object of interest, or as a second number beside the sampled one.
The floor and action-rule checks apply equally to held-out return reported on its own, as most works report it: miner's argmax falls below the floor on held-out return directly, with no gap involved.
Sampling is not self-interpreting either: heist's sampled numbers are accurate, and only the floor shows that they are consistent with chance.

\textbf{What a trustworthy gap requires.} Evaluate a fixed checkpoint; state the action rule and apply it identically to both level sets; seed every random draw in the evaluation pipeline, including the level-subset draw; measure the floor on both level sets with the same harness and draws; specify the statistical tests before the analysis and correct for the number of environments; and read an entropy flag against the above-floor check.
Each step is cheap, and the floor costs minutes.

\textbf{Reliability, stated precisely.} Our claim concerns whether a number supports the inference drawn from it, not that such numbers are noise, and not that a high-entropy policy is illegitimate.
An environment whose held-out return sits well above the floor (miner, \N{miner.8M.s.test} sampled against \N{floor.miner.test}, in the intermediate tier) performs well above chance, and its gap is interpretable whatever its entropy tier; a policy not distinguishable from the floor (heist) supports much less.

\section{Limitations}
\label{sec:limitations}

\textbf{Competence of the agents.} On the easy-mode scale of \citet{cobbe2020procgen} (Appendix~\ref{sec:app_cobbe}), our 8M agents reach a mean normalized held-out score of \N{cobbe.mean.8M.sampled_test} (training \N{cobbe.mean.8M.sampled_train}), below what standard 25M training reaches. Read against their easy-mode curves at the same step count (their Figure~13; approximate, and evaluated on the full level distribution), starpilot, fruitbot, bigfish and heist are in a similar range, miner slightly lower, dodgeball lower on training levels, and coinrun and bossfight clearly lower. Two configuration choices likely contribute: our trainer applies no reward normalization or clipping, unlike CleanRL and \citet{cobbe2020procgen}, and the 8M runs use 16 parallel environments. The findings describe checkpoints at this budget and configuration, the regime of compute-limited studies and early checkpoints. At 25M, where returns are higher (mean normalized held-out score \N{cobbe.mean.25M.sampled_test} over three games), the floor findings persist: heist's greedy held-out return remains below the floor and its sampled one is not distinguishable from it; their own easy-mode heist held-out curve ends near 2 (read from their figure, under their protocol), close to the random floor we measure, so the absence of above-chance transfer on heist is not specific to our configuration.

\textbf{Seed counts and power.} Most conditions use three seeds per encoder (six pooled runs; 25M: nine per encoder for starpilot, six for heist).
With six pooled runs and Holm correction across eight environments, moderate above-floor effects that differ by encoder, as in bigfish and bossfight, are not resolved.
The encoder comparison turns on seed count, and we present it as a case study.

\textbf{The floor is one reference.} A uniform-random policy calibrates only the lower end, and weakly: a trivially achievable reference, such as the best constant action or a policy trained without observations (the $R_{\min}$ of \citet{cobbe2020procgen}), would be stricter. On that scale dodgeball, coinrun, bigfish and bossfight sit near zero (normalized held-out scores \N{cobbe.dodgeball.8M.sampled_test}, \N{cobbe.coinrun.8M.sampled_test}, \N{cobbe.bigfish.8M.sampled_test} and \N{cobbe.bossfight.8M.sampled_test}; Appendix~\ref{sec:app_cobbe}), so for them ``above the floor'' means above chance, not evidence of competent transfer. A scripted or strongly trained reference would calibrate the upper end, and we do not report one.
Random-weight guessing \citep{oller2020rwg} would give a structured lower reference that we have not measured.

\textbf{Action rules in our tables.} All tables report sampled evaluation as primary and greedy beside it; both come from the same seeded protocol, so the greedy columns show what argmax evaluation would have reported for the same checkpoints and levels.

\textbf{One threshold, one suite, one algorithm.} The screen uses a single set of cuts, swept but not derived, on eight of ProcGen's sixteen games in easy mode, with one algorithm (PPO), over raw rather than functionally distinct actions.
The equivalence classes are measured from the rendered observation over a five-step horizon and could miss later effects.

\textbf{Parallel environments.} The 8M runs use 16 parallel environments and the 25M runs 256, against the CleanRL default of 64, so the 8M-to-25M comparison changes batch size and the number of updates along with the step budget.
\citet{beukman2026stagnation} show that PPO stagnation can be relieved by scaling parallel environments; the 8M entropy levels may partly reflect the smaller batch.
The action-rule and floor findings describe the checkpoints as trained and do not depend on why a policy has high residual entropy; the 8M-to-25M comparisons do.

\textbf{Encoders.} Both encoders are trained from scratch; we do not test pretrained encoders, or more recent designs such as Impoola's globally pooled IMPALA encoder \citep{trumpp2025impoola}.

\textbf{Audit scope.} The audit covers released evaluation code at pinned commits, not reported results; papers may have used unreleased evaluation scripts.

\section{Conclusion}
\label{sec:conclusion}

A generalization gap is a measurement, and without a reference point it is hard to say what it measures.
In ProcGen, a measured random floor shows that the test-time action rule decides which policy gets measured (a policy that scores \N{ratio.miner.8M.s.test}$\times$ chance on held-out levels is scored below chance by its own argmax) and that a raw-entropy screen over-calls: of the \N{count.raw.8M.word} environments it flags, four have sampled policies clearly above the floor, bossfight is inconclusive at six runs, and heist's held-out return is not distinguishable from chance.
The field's policy-gradient evaluation code samples actions, in nine of eleven codebases with no explicit choice at the evaluation call site.
We recommend that every reported gap state its action rule, seed its evaluation protocol and specify its tests before the analysis, and report the random floor on both level sets.

\bibliographystyle{tmlr}
\bibliography{refs}

\appendix

\section{Pooled Significance of the Gap}
\label{sec:app_pooling}

The primary summary of a gap is its mean over runs with a 95\% percentile bootstrap CI over runs ($10^4$ resamples). With six runs the bootstrap distribution is coarse and can understate uncertainty, so intervals that end near zero should be read cautiously. As secondary tests, each independently trained run (one seed and encoder) is one draw for pooling: a Welch $t$-test compares its 384 training-level and 384 held-out episode returns, and per-run two-sided $p$-values are combined with Fisher's method, with Stouffer's $Z$ on one-sided (train $>$ test) $p$-values as a directional check.
Runs use disjoint RNG streams, so the combination is valid across runs; the three draws within a run are pooled before testing, not combined.
Holm correction is applied within each family (rule $\times$ budget).
Fisher is sensitive to any run showing a difference and Stouffer additionally requires agreement in direction; the bootstrap CI is on the mean of per-run gaps, so the three can disagree.

\begin{table}[htbp]
  \caption{Generalization gap (train $-$ test) at both budgets for the three extended environments, sampled (primary) and greedy.}
  \label{tab:pooled}
  \centering
  \small
  \setlength{\tabcolsep}{3pt}
  \begin{tabular}{llrlrrrl}
\toprule
Env & Budget & $n$ & Sampled gap [95\% CI] & Fisher $p$ & Holm $p$ & Stouffer $Z$ & Greedy gap [95\% CI] \\
\midrule
starpilot & 8M & 6 & $-0.23$ $[-0.79, 0.48]$ & $0.82$ & $1.00$ & $-0.45$ & $-1.26$ $[-2.71, 0.20]$ \\
starpilot & 25M & 18 & $+2.11$ $[1.39, 2.77]$ & $1.0\times10^{-7}$ & $1.0\times10^{-7}$ & $6.46$ & $+2.29$ $[1.13, 3.50]$ \\
fruitbot & 8M & 6 & $+1.22$ $[0.51, 1.92]$ & $4.5\times10^{-4}$ & $0.002$ & $3.98$ & $+0.88$ $[0.05, 1.75]$ \\
fruitbot & 25M & 6 & $+5.92$ $[5.28, 6.70]$ & $<10^{-30}$ & $<10^{-30}$ & $19.53$ & $+5.46$ $[4.60, 6.19]$ \\
heist & 8M & 6 & $+1.54$ $[1.26, 1.82]$ & $2.3\times10^{-25}$ & $1.9\times10^{-24}$ & $10.95$ & $+0.71$ $[0.44, 0.98]$ \\
heist & 25M & 12 & $+3.94$ $[3.61, 4.30]$ & $<10^{-30}$ & $<10^{-30}$ & $39.74$ & $+2.62$ $[2.14, 3.10]$ \\
\bottomrule
\end{tabular}

\end{table}

\begin{table}[htbp]
  \caption{Generalization gap at 8M for all eight environments.}
  \label{tab:gap8m}
  \centering
  \scriptsize
  \setlength{\tabcolsep}{6pt}
  \begin{tabular}{lrlrrrlrr}
\toprule
Env & $n$ & Sampled gap [95\% CI] & Fisher $p$ & Holm $p$ & $Z$ & Greedy gap [95\% CI] & Fisher $p$ & Holm $p$ \\
\midrule
starpilot & 6 & $-0.23$ $[-0.79, 0.48]$ & $0.82$ & $1.00$ & $-0.45$ & $-1.26$ $[-2.71, 0.20]$ & $0.001$ & $0.003$ \\
fruitbot & 6 & $+1.22$ $[0.51, 1.92]$ & $4.5\times10^{-4}$ & $0.002$ & $3.98$ & $+0.88$ $[0.05, 1.75]$ & $8.7\times10^{-5}$ & $2.6\times10^{-4}$ \\
bigfish & 6 & $+1.58$ $[0.92, 2.42]$ & $6.9\times10^{-23}$ & $4.2\times10^{-22}$ & $10.22$ & $+1.43$ $[0.70, 2.25]$ & $1.3\times10^{-19}$ & $6.3\times10^{-19}$ \\
coinrun & 6 & $+0.02$ $[-0.22, 0.23]$ & $0.70$ & $1.00$ & $0.14$ & $+0.19$ $[-0.03, 0.40]$ & $0.33$ & $0.33$ \\
miner & 6 & $+1.55$ $[1.05, 1.97]$ & $3.2\times10^{-20}$ & $1.6\times10^{-19}$ & $9.50$ & $+0.61$ $[0.31, 0.92]$ & $2.0\times10^{-24}$ & $1.6\times10^{-23}$ \\
dodgeball & 6 & $+0.59$ $[0.37, 0.78]$ & $6.9\times10^{-25}$ & $4.8\times10^{-24}$ & $10.16$ & $+0.46$ $[0.18, 0.74]$ & $1.3\times10^{-22}$ & $9.2\times10^{-22}$ \\
bossfight & 6 & $+0.34$ $[0.04, 0.70]$ & $0.012$ & $0.035$ & $2.61$ & $+0.49$ $[-0.11, 1.09]$ & $2.5\times10^{-7}$ & $1.0\times10^{-6}$ \\
heist & 6 & $+1.54$ $[1.26, 1.82]$ & $2.3\times10^{-25}$ & $1.9\times10^{-24}$ & $10.95$ & $+0.71$ $[0.44, 0.98]$ & $6.9\times10^{-21}$ & $4.1\times10^{-20}$ \\
\bottomrule
\end{tabular}

\end{table}

\FloatBarrier
\section{Floor on Both Level Sets}
\label{sec:app_floor}

Table~\ref{tab:floor} gives the floor and the pooled returns on both level sets at both budgets, on the same level draws, with the primary calls of Table~\ref{tab:unmask}.

\begin{table}[H]
  \caption{Uniform-random floor and pooled returns on both level sets and both budgets, same draws. Floor $\pm$ per-episode SEM (384 episodes); returns $\pm$ SEM over runs. Calls as in Table~\ref{tab:unmask}.}
  \label{tab:floor}
  \centering
  \scriptsize
  \setlength{\tabcolsep}{3pt}
  \begin{tabular}{llrrrll}
\toprule
Env & Levels & Floor & Sampled & Greedy & Sampled call & Greedy call \\
\midrule
starpilot (8M) & train & $1.40 \pm 0.09$ & $18.12 \pm 0.92$ & $16.64 \pm 1.22$ & above & above \\
 & test & $1.67 \pm 0.11$ & $18.35 \pm 1.11$ & $17.89 \pm 1.40$ & above & above \\
fruitbot (8M) & train & $-2.43 \pm 0.20$ & $24.10 \pm 0.44$ & $26.68 \pm 0.27$ & above & above \\
 & test & $-2.66 \pm 0.21$ & $22.89 \pm 0.39$ & $25.79 \pm 0.41$ & above & above \\
bigfish (8M) & train & $0.90 \pm 0.07$ & $4.40 \pm 1.04$ & $4.62 \pm 1.19$ & n.d. & n.d. \\
 & test & $0.84 \pm 0.06$ & $2.82 \pm 0.75$ & $3.19 \pm 0.89$ & n.d. & n.d. \\
coinrun (8M) & train & $2.94 \pm 0.23$ & $5.74 \pm 0.36$ & $3.76 \pm 0.62$ & above & n.d. \\
 & test & $3.28 \pm 0.24$ & $5.72 \pm 0.31$ & $3.57 \pm 0.62$ & above & n.d. \\
miner (8M) & train & $1.24 \pm 0.10$ & $7.48 \pm 0.32$ & $1.21 \pm 0.21$ & above & n.d. \\
 & test & $1.17 \pm 0.10$ & $5.93 \pm 0.30$ & $0.60 \pm 0.07$ & above & below \\
dodgeball (8M) & train & $0.57 \pm 0.06$ & $1.69 \pm 0.18$ & $1.27 \pm 0.15$ & above & above \\
 & test & $0.47 \pm 0.05$ & $1.09 \pm 0.13$ & $0.82 \pm 0.05$ & above & above \\
bossfight (8M) & train & $0.01 \pm 0.01$ & $1.70 \pm 0.63$ & $2.77 \pm 0.96$ & n.d. & n.d. \\
 & test & $0.09 \pm 0.05$ & $1.36 \pm 0.49$ & $2.28 \pm 0.69$ & n.d. & n.d. \\
heist (8M) & train & $3.54 \pm 0.24$ & $4.42 \pm 0.29$ & $1.16 \pm 0.14$ & n.d. & below \\
 & test & $3.15 \pm 0.24$ & $2.88 \pm 0.21$ & $0.45 \pm 0.14$ & n.d. & below \\
\midrule
starpilot (25M) & train & $1.40 \pm 0.09$ & $29.98 \pm 1.02$ & $31.91 \pm 1.31$ & above & above \\
 & test & $1.67 \pm 0.11$ & $27.87 \pm 1.13$ & $29.62 \pm 1.36$ & above & above \\
fruitbot (25M) & train & $-2.43 \pm 0.20$ & $26.88 \pm 0.38$ & $28.25 \pm 0.35$ & above & above \\
 & test & $-2.66 \pm 0.21$ & $20.96 \pm 0.38$ & $22.79 \pm 0.56$ & above & above \\
heist (25M) & train & $3.54 \pm 0.24$ & $6.94 \pm 0.14$ & $3.55 \pm 0.23$ & above & n.d. \\
 & test & $3.15 \pm 0.24$ & $3.00 \pm 0.20$ & $0.92 \pm 0.13$ & n.d. & below \\
\bottomrule
\end{tabular}

\end{table}

\clearpage
\section{Threshold Sensitivity of the Screen}
\label{sec:app_threshold}

Table~\ref{tab:sweep} moves the low/intermediate cut across $[1.8, 2.2]$ and reports each environment's tier under raw and per-state merged entropy.

\begin{table}[H]
  \caption{Tier under a swept low/intermediate cut $T$ (high cut fixed at 2.3; merged cuts at $T/\ln 15$ and $2.3/\ln 15$ of each state's maximum). L/I/H = low/intermediate/high residual entropy. 8M, sampled-evaluation states.}
  \label{tab:sweep}
  \centering
  \small
  \begin{tabular}{lrcccccccccc}
\toprule
 & \multicolumn{6}{c}{Raw} & \multicolumn{5}{c}{Merged (per state)} \\
\cmidrule(lr){2-7}\cmidrule(lr){8-12}
Env & $H$ & $1.8$ & $1.9$ & $2.0$ & $2.1$ & $2.2$ & $1.8$ & $1.9$ & $2.0$ & $2.1$ & $2.2$ \\
\midrule
starpilot & $1.794$ & L & L & L & L & L & L & L & L & L & L \\
fruitbot & $2.078$ & I & I & I & L & L & L & L & L & L & L \\
bigfish & $1.971$ & I & I & L & L & L & L & L & L & L & L \\
coinrun & $2.261$ & I & I & I & I & I & I & I & I & L & L \\
miner & $2.087$ & I & I & I & L & L & L & L & L & L & L \\
dodgeball & $2.427$ & H & H & H & H & H & I & I & L & L & L \\
bossfight & $2.329$ & H & H & H & H & H & I & L & L & L & L \\
heist & $2.569$ & H & H & H & H & H & I & I & I & I & L \\
\bottomrule
\end{tabular}

\end{table}

\FloatBarrier
\section{Calibration: Effective Actions and Merged Entropy}
\label{sec:app_calib}

To test whether the 15 actions are equally meaningful, we sampled 400 states per game by random play, cloned each state, applied every action from the identical state, and compared the resulting $64{\times}64$ observation, reward and episode boundary over the action and four subsequent no-ops.
Replaying the same action from the same state gave identical outcomes in every case, so equal outcomes identify equivalent actions.
The number of distinct outcome classes across sampled states ($k$, Table~\ref{tab:merged}) ranges from 4 (fruitbot) to 11 (starpilot); the special-action keys are inert in most games, and in fruitbot and miner the three leftward and three rightward moves each collapse to one.
For merged entropy we sum the policy's probabilities within each class, either using the classes of the state being scored (found by the same cloning procedure on a seeded uniform sample of 1024 evaluation states per cell) or using one partition per game (actions equivalent at every sampled state).
Normalizing raw entropy by $\ln k$ is not meaningful, because raw entropy exceeds $\ln k$ where mass is spread over equivalent actions; merging first is the valid comparison.

\begin{table}[htbp]
  \caption{Entropy (nats) on sampled-evaluation states at 8M (upper block) and 25M (lower block) before and after merging functionally equivalent actions; share is the fraction of raw entropy within classes, measured on the same states. Last two columns: raw entropy on greedy-evaluation states, and on the first 1024 sampled-evaluation states only.}
  \label{tab:merged}
  \centering
  \small
  \begin{tabular}{lrrrrrrr}
\toprule
Env & $k$ (global) & Raw $H$ & Merged $H$ (state) & Merged $H$ (game) & Share & Greedy states & First 1024 \\
\midrule
starpilot & 11 & $1.794$ & $1.124$ & $1.217$ & $37\%$ & $1.851$ & $1.785$ \\
fruitbot & 4 & $2.077$ & $0.697$ & $0.791$ & $66\%$ & $2.123$ & $2.006$ \\
bigfish & 9 & $1.974$ & $1.321$ & $1.404$ & $33\%$ & $2.011$ & $2.088$ \\
coinrun & 9 & $2.261$ & $0.928$ & $1.523$ & $59\%$ & $2.263$ & $2.030$ \\
miner & 5 & $2.083$ & $0.716$ & $1.109$ & $66\%$ & $1.897$ & $1.726$ \\
dodgeball & 10 & $2.426$ & $1.655$ & $1.798$ & $32\%$ & $2.440$ & $2.270$ \\
bossfight & 10 & $2.328$ & $1.366$ & $1.715$ & $41\%$ & $2.357$ & $2.446$ \\
heist & 9 & $2.569$ & $1.689$ & $1.782$ & $34\%$ & $2.546$ & $2.519$ \\
\midrule
starpilot (25M) & 11 & $1.583$ & $0.898$ & $0.970$ & $43\%$ & $1.665$ & $2.145$ \\
fruitbot (25M) & 4 & $2.220$ & $0.722$ & $0.811$ & $67\%$ & $2.274$ & $2.195$ \\
heist (25M) & 9 & $2.455$ & $1.640$ & $1.740$ & $33\%$ & $2.360$ & $2.224$ \\
\bottomrule
\end{tabular}

\end{table}

\section{Exploratory: Merged-Argmax Evaluation}
\label{sec:app_merged}

This analysis was not part of the analysis plan of Section~\ref{sec:stats}. Its rule, tests and interpretation were written down before any cell was run. The 48 checkpoints at 8M were evaluated on both level sets and the same three seeded draws as Table~\ref{tab:unmask}, with a deterministic rule: sum the softmax probabilities within each per-game equivalence class (Appendix~\ref{sec:app_calib}), take the class with the largest sum, then the most probable raw action inside it. Floor calls use the tests of Section~\ref{sec:stats} with Holm correction within this analysis only. The disagreement rate is the fraction of counted evaluation steps on which the merged choice falls in a different class from the plain argmax action at the same state.

\begin{table}[H]
  \caption{Exploratory: held-out return under merged argmax at 8M, against plain argmax (greedy), sampling and the floor on the same draws (mean $\pm$ SEM over six runs). Merged call: Holm within this analysis. Merged$-$greedy: paired difference over runs. Disagree.: share of steps on which merged argmax picks a different class from plain argmax.}
  \label{tab:merged_argmax}
  \centering
  \small
  \setlength{\tabcolsep}{4pt}
  \begin{tabular}{lrrrrlrr}
\toprule
Env & Floor & Greedy & Merged & Sampled & Merged call & Merged$-$greedy & Disagree. \\
\midrule
starpilot & $1.67$ & $17.89 \pm 1.40$ & $14.70 \pm 1.11$ & $18.35 \pm 1.11$ & above & $-3.20$ & $33\%$ \\
fruitbot & $-2.66$ & $25.79 \pm 0.41$ & $26.07 \pm 0.67$ & $22.89 \pm 0.39$ & above & $+0.28$ & $21\%$ \\
bigfish & $0.84$ & $3.19 \pm 0.89$ & $2.67 \pm 0.71$ & $2.82 \pm 0.75$ & n.d. & $-0.52$ & $43\%$ \\
coinrun & $3.28$ & $3.57 \pm 0.62$ & $2.71 \pm 0.59$ & $5.72 \pm 0.31$ & n.d. & $-0.86$ & $61\%$ \\
miner & $1.17$ & $0.60 \pm 0.07$ & $0.47 \pm 0.07$ & $5.93 \pm 0.30$ & below & $-0.13$ & $38\%$ \\
dodgeball & $0.47$ & $0.82 \pm 0.05$ & $0.13 \pm 0.05$ & $1.09 \pm 0.13$ & below & $-0.69$ & $68\%$ \\
bossfight & $0.09$ & $2.28 \pm 0.69$ & $1.18 \pm 0.65$ & $1.36 \pm 0.49$ & n.d. & $-1.10$ & $73\%$ \\
heist & $3.15$ & $0.45 \pm 0.14$ & $0.08 \pm 0.08$ & $2.88 \pm 0.21$ & below & $-0.37$ & $73\%$ \\
\bottomrule
\end{tabular}

\end{table}

\FloatBarrier

\section{Normalized Scores on the Scale of Cobbe et al.}
\label{sec:app_cobbe}

Table~\ref{tab:cobbe} reports pooled returns as $(R - R_{\min})/(R_{\max} - R_{\min})$ with the easy-mode constants of \citet{cobbe2020procgen} (their Appendix~C). Their $R_{\min}$ comes from a policy trained with observations masked out, which is why the measured uniform-random floor is usually negative on this scale.

\begin{table}[H]
  \caption{Normalized scores (mean over runs; each run averaged over three seeded draws). Sampled is primary. The floor is the uniform-random policy on the same draws.}
  \label{tab:cobbe}
  \centering
  \small
  \setlength{\tabcolsep}{4pt}
  \begin{tabular}{lrrrrrrr}
\toprule
 & & & \multicolumn{2}{c}{Sampled} & \multicolumn{1}{c}{Greedy} & \multicolumn{2}{c}{Uniform floor} \\
\cmidrule(lr){4-5}\cmidrule(lr){6-6}\cmidrule(lr){7-8}
Env & Budget & $R_{\min}$, $R_{\max}$ & Train & Test & Test & Train & Test \\
\midrule
starpilot & 8M & 2.5, 64 & 0.25 & 0.26 & 0.25 & $-$0.02 & $-$0.01 \\
fruitbot & 8M & -1.5, 32.4 & 0.76 & 0.72 & 0.81 & $-$0.03 & $-$0.03 \\
bigfish & 8M & 1, 40 & 0.09 & 0.05 & 0.06 & $-$0.00 & $-$0.00 \\
coinrun & 8M & 5, 10 & 0.15 & 0.14 & $-$0.29 & $-$0.41 & $-$0.34 \\
miner & 8M & 1.5, 13 & 0.52 & 0.39 & $-$0.08 & $-$0.02 & $-$0.03 \\
dodgeball & 8M & 1.5, 19 & 0.01 & $-$0.02 & $-$0.04 & $-$0.05 & $-$0.06 \\
bossfight & 8M & 0.5, 13 & 0.10 & 0.07 & 0.14 & $-$0.04 & $-$0.03 \\
heist & 8M & 3.5, 10 & 0.14 & $-$0.10 & $-$0.47 & 0.01 & $-$0.05 \\
\midrule
starpilot & 25M & 2.5, 64 & 0.45 & 0.41 & 0.44 & $-$0.02 & $-$0.01 \\
fruitbot & 25M & -1.5, 32.4 & 0.84 & 0.66 & 0.72 & $-$0.03 & $-$0.03 \\
heist & 25M & 3.5, 10 & 0.53 & $-$0.08 & $-$0.40 & 0.01 & $-$0.05 \\
\midrule
mean & 8M & & 0.25 & 0.19 & 0.05 & & $-$0.07 \\
mean & 25M & & 0.60 & 0.33 & 0.25 & & $-$0.03 \\
\bottomrule
\end{tabular}

\end{table}

\FloatBarrier

\section{Audit: Commit Hashes and Evidence Locations}
\label{sec:app_audit}

Table~\ref{tab:commits} lists, for each audited codebase, the source commit examined and the file and line evidencing each property.
Where an action-selection call takes a \texttt{deterministic} argument, the default-\texttt{False} definition site is noted.
For EDE's own value-based agent, \texttt{train\_rainbow.py:642} applies an evaluation $\epsilon$ whose default is set at \texttt{level\_replay/dqn\_args.py:101}.

\begin{table}[htbp]
  \caption{Commit hash and evidence location per codebase. S = stochastic test evaluation, L = in-loop test return. For L rows the location is where the held-out environment or worker is created, then where its running return is reported.}
  \label{tab:commits}
  \centering
  \scriptsize
  \setlength{\tabcolsep}{3pt}
  \begin{tabular}{lllccl}
    \toprule
    Codebase & Venue & Commit & S & L & Location (file:line) \\
    \midrule
    UCB-DrAC & NeurIPS'21 & \texttt{1fbf2567cca1} & \checkmark & -- & \texttt{test.py:42} (\texttt{deterministic=False}) \\
    DAAC     & ICML'21    & \texttt{2fe302029428} & \checkmark & -- & \texttt{test.py:31} (\texttt{act()}, default False) \\
    PLR      & ICML'21    & \texttt{ccecf452ee33} & \checkmark & -- & \texttt{test.py:71} (default False, \texttt{test.py:34}) \\
    EDE      & NeurIPS'23 & \texttt{89396f548aaa} & \checkmark & -- & \texttt{test.py:70} (PPO; default \texttt{:30}); \texttt{train\_rainbow.py:642} ($\epsilon$) \\
    Impoola  & RLC'25     & \texttt{1d71110451cc} & \checkmark & -- & \texttt{evaluation.py:30}; \texttt{ppo\_training.py:47} (F), \texttt{dqn\_training.py:50} (T) \\
    VSOP     & ICML'24    & \texttt{7764611bdc98} & \checkmark & \checkmark & \texttt{ppo\_procgen.py:368} (\texttt{probs.sample()}); \texttt{:457} $\to$ \texttt{:691} \\
    VSOP-3D  & arXiv'24   & \texttt{6eb3ce463b47} & \checkmark & \checkmark & \texttt{vsop\_3d\_procgen.py:354} (\texttt{probs.sample()}); \texttt{:440} $\to$ \texttt{:693} \\
    mixreg   & NeurIPS'20 & \texttt{0800f492491f} & \checkmark & \checkmark & \texttt{train.py:66--72} test worker $\to$ \texttt{ppo2.py:219} \\
    RAD      & NeurIPS'20 & \texttt{d61a1ac8258a} & \checkmark & \checkmark & \texttt{train.py:77} eval venv $\to$ \texttt{ppo2.py:220} \\
    IBAC-SNI & NeurIPS'19 & \texttt{6b3a58bfc23a} & \checkmark & \checkmark & \texttt{policies.py:196} (\texttt{pd\_run.sample()}); \texttt{config.py:198} $\to$ \texttt{ppo2.py:395} \\
    ProcGen  & ICML'20    & \texttt{1a2ae2194a61} & \checkmark & \checkmark & \texttt{train.py:28} test worker $\to$ baselines \texttt{ppo2.py:201} \\
    PPG      & ICML'21    & \texttt{7295473f0185} & n/a & n/a & no held-out evaluation \\
    \bottomrule
  \end{tabular}
\end{table}

Repositories on GitHub, given as owner/repository (third-party code, at the commits above): UCB-DrAC: \texttt{rraileanu/auto-drac}; DAAC: \texttt{rraileanu/idaac}; PLR: \texttt{facebookresearch/level-replay}; EDE: \texttt{facebookresearch/ede}; Impoola: \texttt{raphajaner/impoola}; VSOP: \texttt{anndvision/vsop}; VSOP-3D: \texttt{anndvision/vsop-3d}; mixreg: \texttt{kaixin96/mixreg}; RAD: \texttt{pokaxpoka/rad\_procgen}; IBAC-SNI: \texttt{microsoft/IBAC-SNI}; ProcGen: \texttt{openai/train-procgen}; PPG: \texttt{openai/phasic-policy-gradient}.

\section{Additional Details and Tables}
\label{sec:app_extra}

\textbf{Encoder details.} Both encoders share a three-block IMPALA backbone (channel widths $[16,32,32]$; each block is a $3{\times}3$ convolution, stride-2 max-pool and two residual sub-blocks) on $(3,64,64)$ RGB input.
The baseline flattens the final $32{\times}8{\times}8$ map and projects to 256 dimensions.
The FPN variant projects each block's feature map to 32 channels with $1{\times}1$ lateral convolutions, fuses them top-down with nearest-neighbor upsampling \citep{lin2017fpn}, smooths each fused map with a $3{\times}3$ convolution, pools to $4{\times}4$, concatenates (1536 dimensions) and projects to 256.
The FPN encoder has fewer parameters (521k against 622k).

\textbf{Evaluation devices.} Of the 1{,}008 evaluation cells, 93 (47 of them sampled) ran on CPU because of memory limits. Sampled actions in those cells come from the CPU random-number stream, so they are reproducible on CPU but not bit-identical to a GPU run with the same seed; greedy cells draw no random numbers.

\begin{table}[htbp]
  \caption{PPO hyperparameters, fixed across encoders and environments (CleanRL ProcGen hyperparameters, except the number of parallel environments, whose default is 64, and the absence of reward normalization and reward clipping).}
  \label{tab:hyper}
  \centering
  \small
  \begin{tabular}{ll}
    \toprule
    Hyperparameter & Value \\
    \midrule
    Parallel environments & 16 (8M); 256 (25M) \\
    Rollout length & 256 \\
    Total timesteps & 8M; 25M \\
    Learning rate & $5\times10^{-4}$ (no annealing) \\
    Discount factor $\gamma$ & 0.999 \\
    GAE $\lambda$ & 0.95 \\
    Minibatches per update & 8 \\
    Update epochs & 3 \\
    Clip coefficient & 0.2 \\
    Value loss clipping & enabled \\
    Entropy coefficient & 0.01 \\
    Value loss coefficient & 0.5 \\
    Max gradient norm & 0.5 \\
    Reward normalization / clipping & none (CleanRL: normalized, clipped to $\pm 10$) \\
    \bottomrule
  \end{tabular}
\end{table}

\begin{table}[htbp]
  \caption{Final-checkpoint results at 8M per encoder, mean $\pm$ SEM over three seeds, each run averaged over three seeded draws.}
  \label{tab:full8m}
  \centering
  \scriptsize
  \setlength{\tabcolsep}{3pt}
  \begin{tabular}{llrrrrrr}
\toprule
 & & \multicolumn{3}{c}{Sampled} & \multicolumn{3}{c}{Greedy} \\
\cmidrule(lr){3-5}\cmidrule(lr){6-8}
Env & Encoder & Train & Test & Gap & Train & Test & Gap \\
\midrule
starpilot & IMPALA & $17.40 \pm 0.30$ & $17.23 \pm 0.83$ & $0.17 \pm 0.64$ & $16.57 \pm 1.35$ & $17.22 \pm 0.82$ & $-0.64 \pm 1.52$ \\
starpilot & FPN & $18.85 \pm 1.90$ & $19.47 \pm 2.05$ & $-0.62 \pm 0.27$ & $16.70 \pm 2.37$ & $18.57 \pm 2.94$ & $-1.87 \pm 0.80$ \\
fruitbot & IMPALA & $24.38 \pm 0.08$ & $22.76 \pm 0.49$ & $1.61 \pm 0.54$ & $26.81 \pm 0.47$ & $25.72 \pm 0.67$ & $1.09 \pm 1.02$ \\
fruitbot & FPN & $23.83 \pm 0.95$ & $23.01 \pm 0.70$ & $0.82 \pm 0.58$ & $26.55 \pm 0.37$ & $25.86 \pm 0.62$ & $0.68 \pm 0.28$ \\
bigfish & IMPALA & $2.10 \pm 0.18$ & $1.22 \pm 0.15$ & $0.88 \pm 0.07$ & $2.10 \pm 0.15$ & $1.25 \pm 0.07$ & $0.85 \pm 0.07$ \\
bigfish & FPN & $6.70 \pm 0.27$ & $4.42 \pm 0.49$ & $2.29 \pm 0.66$ & $7.14 \pm 0.85$ & $5.14 \pm 0.38$ & $2.01 \pm 0.79$ \\
coinrun & IMPALA & $6.51 \pm 0.19$ & $6.35 \pm 0.26$ & $0.16 \pm 0.17$ & $5.07 \pm 0.31$ & $4.81 \pm 0.18$ & $0.26 \pm 0.14$ \\
coinrun & FPN & $4.97 \pm 0.13$ & $5.10 \pm 0.05$ & $-0.12 \pm 0.18$ & $2.45 \pm 0.36$ & $2.34 \pm 0.59$ & $0.11 \pm 0.23$ \\
miner & IMPALA & $7.63 \pm 0.45$ & $5.60 \pm 0.35$ & $2.03 \pm 0.15$ & $1.65 \pm 0.15$ & $0.70 \pm 0.12$ & $0.95 \pm 0.18$ \\
miner & FPN & $7.33 \pm 0.53$ & $6.26 \pm 0.48$ & $1.06 \pm 0.28$ & $0.77 \pm 0.11$ & $0.50 \pm 0.05$ & $0.27 \pm 0.07$ \\
dodgeball & IMPALA & $1.40 \pm 0.16$ & $0.93 \pm 0.06$ & $0.47 \pm 0.21$ & $1.13 \pm 0.28$ & $0.83 \pm 0.04$ & $0.30 \pm 0.27$ \\
dodgeball & FPN & $1.98 \pm 0.22$ & $1.26 \pm 0.25$ & $0.72 \pm 0.06$ & $1.41 \pm 0.12$ & $0.80 \pm 0.10$ & $0.61 \pm 0.16$ \\
bossfight & IMPALA & $0.42 \pm 0.06$ & $0.37 \pm 0.09$ & $0.05 \pm 0.11$ & $0.83 \pm 0.06$ & $1.08 \pm 0.21$ & $-0.26 \pm 0.16$ \\
bossfight & FPN & $2.98 \pm 0.61$ & $2.35 \pm 0.45$ & $0.62 \pm 0.28$ & $4.72 \pm 0.91$ & $3.48 \pm 0.96$ & $1.24 \pm 0.14$ \\
heist & IMPALA & $3.91 \pm 0.38$ & $2.45 \pm 0.11$ & $1.47 \pm 0.29$ & $1.15 \pm 0.23$ & $0.35 \pm 0.09$ & $0.80 \pm 0.28$ \\
heist & FPN & $4.92 \pm 0.18$ & $3.31 \pm 0.12$ & $1.61 \pm 0.20$ & $1.17 \pm 0.23$ & $0.56 \pm 0.28$ & $0.62 \pm 0.19$ \\
\bottomrule
\end{tabular}

\end{table}

\begin{table}[htbp]
  \caption{Heist at both budgets, the same checkpoints under both rules, pooled over encoders (mean $\pm$ SEM over runs), with the uniform-random floor on the identical level draws. $^\dagger$For the floor rows, the Gap column is floor train $-$ floor test.}
  \label{tab:gap2x2}
  \centering
  \small
  \begin{tabular}{llrrr}
\toprule
Budget & Rule & Train & Test & Gap \\
\midrule
8M ($n{=}6$) & sampled & $4.42 \pm 0.29$ & $2.88 \pm 0.21$ & $1.54 \pm 0.16$ \\
8M ($n{=}6$) & greedy & $1.16 \pm 0.14$ & $0.45 \pm 0.14$ & $0.71 \pm 0.16$ \\
 & uniform floor$^\dagger$ & $3.54 \pm 0.24$ & $3.15 \pm 0.24$ & $+0.39$ \\
\midrule
25M ($n{=}12$) & sampled & $6.94 \pm 0.14$ & $3.00 \pm 0.20$ & $3.94 \pm 0.19$ \\
25M ($n{=}12$) & greedy & $3.55 \pm 0.23$ & $0.92 \pm 0.13$ & $2.62 \pm 0.26$ \\
 & uniform floor$^\dagger$ & $3.54 \pm 0.24$ & $3.15 \pm 0.24$ & $+0.39$ \\
\bottomrule
\end{tabular}

\end{table}

\FloatBarrier

\section{Representation Diagnostics}
\label{sec:app_cka}

Linear centered kernel alignment (CKA) \citep{kornblith2019cka, davari2022cka} between two representation matrices is a sum over paired rows, so row $i$ of each must be the same input.
A within-encoder train-versus-test comparison, which sets 1024 training-level observations against 1024 different held-out observations, has no valid pairing and is not a measurement of train-specificity.
Its values sit far above the row-permutation null (seed-mean $z$ from \N{ctl.z.min} to \N{ctl.z.max}), but a $2\times2$ control shows what drives them (Table~\ref{tab:ckacontrol}): comparing train against train at matched timesteps (C) gives the same value as train against test (A), and shuffling one side's rows (B, D) collapses both, by \N{ctl.fruitbot.collapse.min}--\N{ctl.fruitbot.collapse.max}$\times$ on fruitbot, so the statistic tracks episode-timestep alignment rather than the level set.
For a uniform row permutation its expectation is exactly $\sqrt{\mathrm{PR}_X\,\mathrm{PR}_Y}/(n-1)$, with PR the participation ratio, so between-encoder differences track effective rank: IMPALA's raw value on fruitbot at 8M is \N{d4.raw}$\times$ FPN's, but \N{d4.analytic}$\times$ in units of each side's null, and seed by seed \N{d4.analytic.s1}, \N{d4.analytic.s2} and \N{d4.analytic.s3}.
On heist at 25M, changing only the collection geometry moves the within-encoder value by \N{geo16.impala.factor}$\times$ (IMPALA) and \N{geo16.fpn.factor}$\times$ (FPN).
Cross-encoder CKA on identical observations is correctly paired but plays no role in our conclusions.

\begin{table}[htbp]
  \caption{Controls on within-encoder train-versus-test CKA, 8M, mean over 3 seeds. ``As computed'', permutation and analytic nulls at $n{=}1024$; the $2\times2$ block at $n{=}512$: A = timestep-aligned train vs.\ test, C = timestep-aligned train vs.\ train, B and D = the same with one side's rows shuffled (mean of 200 permutations). PR: participation ratio.}
  \label{tab:ckacontrol}
  \centering
  \small
  \setlength{\tabcolsep}{3pt}
  \begin{tabular}{llrrrrrrrr}
\toprule
& & \multicolumn{3}{c}{$n{=}1024$} & \multicolumn{4}{c}{$2\times2$ ($n{=}512$)} & \\
\cmidrule(lr){3-5}\cmidrule(lr){6-9}
Env & Enc. & As computed & Perm. & Analytic & A & C & B & D & PR \\
\midrule
starpilot & IMPALA & 0.020 & 0.001 & 0.001 & 0.008 & 0.013 & 0.003 & 0.002 & 1.4 \\
starpilot & FPN & 0.020 & 0.002 & 0.002 & 0.014 & 0.012 & 0.003 & 0.003 & 1.6 \\
fruitbot & IMPALA & 0.144 & 0.023 & 0.023 & 0.259 & 0.251 & 0.040 & 0.039 & 22.6 \\
fruitbot & FPN & 0.042 & 0.008 & 0.008 & 0.081 & 0.097 & 0.015 & 0.014 & 7.4 \\
\bottomrule
\end{tabular}

\end{table}

\end{document}